# Cast and Self Shadow Segmentation in Video Sequences using Interval based Eigen Value Representation

| Chandrajit M. | Girisha R. | Vasudev T. | Ashok C.B. |
|---|---|---|---|
| Maharaja Research Foundation | PET Research Centre | Maharaja Research Foundation | Maharaja Research Foundation |
| MIT, Mysore | PESCE, Mandya | MIT, Mysore | MIT, Mysore |
| India | India | India | India |

## ABSTRACT
Tracking of motion objects in the surveillance videos is useful for the monitoring and analysis. The performance of the surveillance system will deteriorate when shadows are detected as moving objects. Therefore, shadow detection and elimination usually benefits the next stages. To overcome this issue, a method for detection and elimination of shadows is proposed. This paper presents a method for segmenting moving objects in video sequences based on determining the Euclidian distance between two pixels considering neighborhood values in temporal domain. Further, a method that segments cast and self shadows in video sequences by computing the Eigen values for the neighborhood of each pixel is proposed. The dual-map for cast and self shadow pixels is represented based on the interval of Eigen values. The proposed methods are tested on the benchmark IEEE CHANGE DETECTION 2014 dataset.

## 1. INTRODUCTION
Motion and shadow segmentation are key processes in many computer vision applications like automated video surveillance and traffic video monitoring. The accurate output of both the process is important for the next stages of applications. Motion segmentation is the process of segmenting pixels under motion in each frame of a video. Whereas, shadow segmentation is the process of segmenting shadow pixels which are part of motion objects in the video. The shadows can be categorized as cast and self shadow as shown in Figure 1. The shadow of the object projected on the background is called cast shadow and the shadow that is projected on the object body itself is called self shadow.

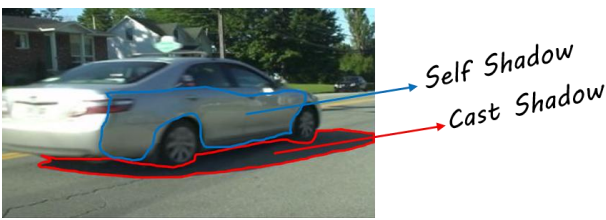

**Fig 1:** Cast and self shadows

Motion segmentation itself is a challenging problem as the surveillance video feed suffers from illumination variations, noise and complex environment [1]. Furthermore, the detection of shadow pixels as cast and self shadow is even more challenging because the shadow pixels will be having same characteristics of the motion object pixels. These issues altogether make motion and shadow segmentation a complex problem [2-4].

To address these issues, this paper presents a method for segmenting motion object, cast and self shadows in video sequences captured in complex environment. The paper is organized as follows: Section 2 reports the literature on motion and shadow segmentation, Section 3 presents the overview of proposed work, section 4 will present the proposed motion segmentation and shadow segmentation algorithm followed by experimental results and conclusions in section 5 and section 6 respectively.

## 2. LITERATURE REVIEW
Most of the research works in motion segmentation have been attempted using the conventional background subtraction [5], statistical background subtraction [6-10], temporal differencing [11-13], optical flow [14] and hybrid [15-21] approaches.

Stauffer and Grimson [10] modeled the background as mixture of Gaussians for motion segmentation. Chao et al. [11] performed moving objects segmentation using frame difference between current and previous frame. Vehicle and person detection using optical flow discontinuities and color information is proposed by Denman et al. [14]. A method that combines temporal differencing with statistical correlation is proposed by Girisha and Murali [17] for motion segmentation. Some more hybrid methods are proposed in [15,16, 18-21] which use combination of motion based and spatio-temporal segmentation for segmenting the objects of interest.

Detection and subsequent elimination of shadows in the frames of video is a vital stage after motion segmentation step. Shadows can cause object merging, object shape distortion and object loss [17]. The information cues used for detecting the shadows are geometry, texture, chromacity, physical and intensity. The shadow detection methods can be categorized as property-based [22-26] and model-based approaches [27-30].

Girisha and Murali [23] proposed a pixel based statistical approach to detect and subsequently eliminate moving cast shadows. They also proposed methods to remove self shadow using statistical difference of mean and ANOVA hypothesis test [24]. Leone and Distante [26] used texture (Gabor filter) to detect cast shadow. A method to detect shadow in a single image based on Eigen analysis is proposed in [27]. Amato et al. [22] detected cast shadow based on luminance ratio of background and foreground pixel values in RGB color space. Jia and Chu [25] used Gaussian Mixture Model to learn the local color features of shadow regions to segment shadow pixels. Nadimi and Bhanu [29] proposed a multistage segmentation based on RGB features to detect shadows. Chia and Aggarwal [28] detected shadows based on extracting features from connected component analysis. Wang et al. [30] used spatial, temporal and edge information in Bayesian network for foreground and shadow detection.

In summary, majority of the work is done on detection and removal of cast shadows. However, a little work is reported towards removal of self shadows but the elimination should be taken into consideration as they create troublesome effect while tracking motion [24].

## 3. OVERVIEW OF THE PROPOSED METHOD

The overview of the proposed work is show in Fig. 2. Motion segmentation for each pixel is done in temporal domain by computing the sum of Euclidean distance between the 3 x 3 pixel neighborhoods of two successive frames with predefined threshold. Hole filling (post processing) step is applied because the blobs extracted from the motion segmentation contains holes due to temporal differencing. Subsequently, for each pixel considering 3 x 3 neighbors Eigen values are computed and the resulting three Eigen values are summed into one. The summed Eigen values are represented in dual-map based on interval values for detection of cast and self shadows. Finally, by superimposing the dual-map on the hole filled frame, cast and self shadows are segmented.

## 4. THE PROPOSED METHOD
### 4.1 Motion Segmentation

In color video frames each pixel will be composed of intensities of red (R), green (G) and blue (B) components. In this paper, the mean of RGB color intensity values of each pixel computed using Eq. 1 is considered as pixel value for motion segmentation and shadow detection.

$$P(w,h) = \frac{P(w,h)_R + P(w,h)_G + P(w,h)_B}{3} \quad (1)$$

where, (w, h) are the dimensions, in which w ranges from 0 to maximum width of frame in horizontal direction and h ranges from 0 to maximum height of frame in vertical direction.

The mean of Euclidean distance between 3 x 3 neighborhood intensity values for each pixel of successive frames is computed using the following equation.

$$\mu_d = \frac{1}{9}\sum_{i=1}^{9} d(P_{f_{t_{N9}}}, P_{f_{t+1_{N9}}}) \quad (2)$$

where, $d$ is the Euclidean distance vector between neighborhood pixel $(P_{f_{t_{N9}}} \& P_{f_{t+1_{N9}}})$ values and $t$ is current temporal frame.

Motion pixels are segmented by comparing the $\mu_d$ with threshold value $T$ and the output frame $D$ which contains moving objects is generated using the following equation.

$$D(w,h) = \begin{cases} RGB \text{ of } I_{P(w,h)} & (\mu_d > T) \\ 0 & (Otherwise) \end{cases} \quad (3)$$

where, $I_{P(w,h)}$ is the pixel in input frame.

A post processing step is employed in order to eliminate any holes in the moving blobs using the method proposed in [31] followed by a morphological operation erosion to remove the remaining noise. Finally, the RGB color intensity values of input frame are superimposed on the frame $D$ to generate $D_{HF}$.

### 4.2 Cast and Self Shadow Segmentation

Eigen values are used in many image processing applications like image compression, image segmentation, image classification and face recognition. Further, the work reported in [27] where, Eigen values is used as a basis to segment cast shadow in image motivated to explore the possibilities of using Eigen values for both cast and self shadow segmentation in video frames. The novelty of the proposed method compared with Souza et al.,[27] is the proposed method aims to segment both the cast and self shadow in the surveillance video sequences using interval value representation of the summed Eigen values.

The shadow segmentation in this paper is based on the analysis of Eigen values computed using Eq. 4 for the 3 x 3 neighborhood of each pixel in the frame $D_{HF}$. Empirical analysis on the computed values of Eq. 4 revealed the cast shadow pixels have low summed Eigen values. Furthermore, the self shadow pixels have high summed Eigen values than cast shadow and object pixels as shown in Fig. 3. This clue was explored and a dual-map representation based on the intervals of summed Eigen values is used for segmentation of cast as well as self shadow pixels.

The Eigen values $\lambda$ are computed for 3 x 3 neighborhood of each pixel in the frame $D_{HF}$.

$$det(D_{HF\,P(w,h)_{N9}} - \lambda I) = 0 \quad (4)$$

where, $I$ is the 3 x 3 Identity matrix.

The summation of Eigen values $\lambda_\Sigma$ for the 3 x 3 neighborhood is computed as:

$$\lambda_\Sigma = \sum_{i=1}^{3} \lambda_i \quad (5)$$

As per the empirical analysis, shadow pixels fall within the minimum and maximum intervals of summed Eigen values as shown in Fig. 3. Therefore, a dual-map scheme based on intervals of summed Eigen values for cast and self shadow pixels is represented as follows:

$$Cast\ Shadow = [\lambda^C_{\Sigma_{min}},\ \lambda^C_{\Sigma_{max}}] \quad (6)$$

$$Self\ Shadow = [\lambda^S_{\Sigma_{min}},\ \lambda^S_{\Sigma_{max}}] \quad (7)$$

where, $(\lambda^C_{\Sigma_{min}},\ \lambda^C_{\Sigma_{max}})$ and $(\lambda^S_{\Sigma_{min}},\ \lambda^S_{\Sigma_{max}})$ are the minimum and maximum summed Eigen Values for cast and self shadow respectively.

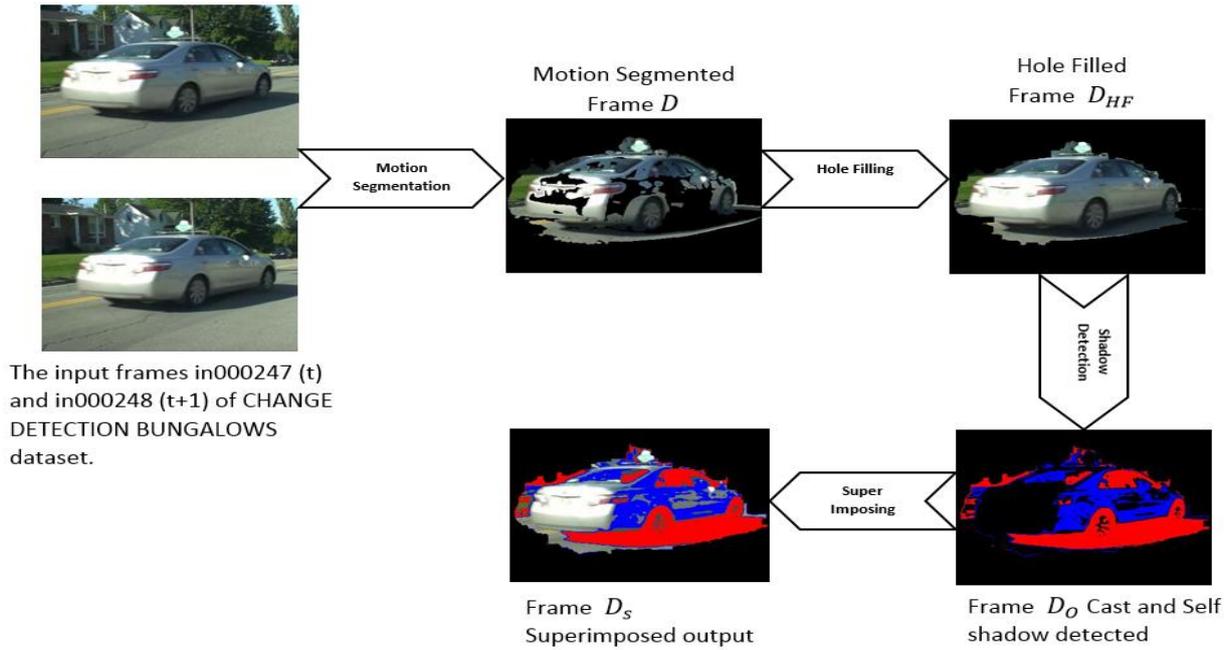

**Fig 1: Block diagram for the proposed method**

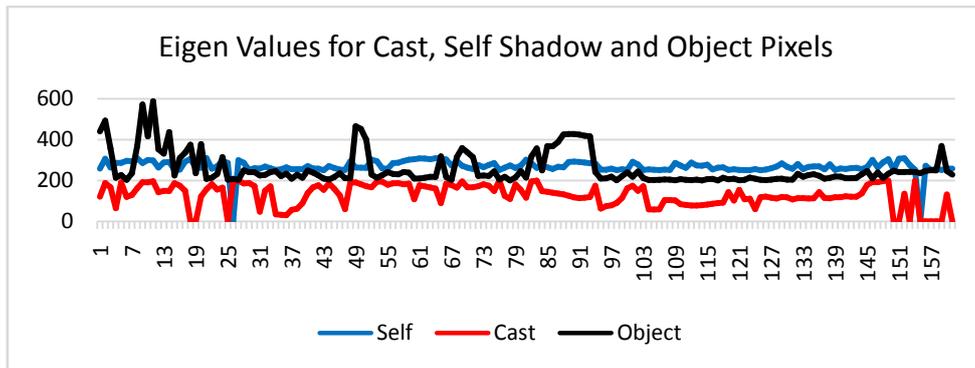

**Fig 3: Summed Eigen values plot for cast, self shadow and object pixels for frame in000136 of IEEE CHANGE DETECTION BUNGALOWS dataset**

The output frame $D_{S_{P(w,h)}}$ in which cast and self shadow pixels are represented as red and blue color intensities based on following equation.

$$D_{S_{P(w,h)}} = \begin{cases} Red & ([\lambda^C_{\Sigma_{min}}, \lambda^C_{\Sigma_{max}}]) \\ Blue & ([\lambda^S_{\Sigma_{min}}, \lambda^S_{\Sigma_{max}}]) \\ RGB\ of\ D_{HF_{P(w,h)}} & (Otherwise, Segmented\ object) \end{cases} \quad (8)$$

## 5. EXPERIMENTS

The proposed work has been implemented using Intel Dual Core 1.8GHz machine with Windows 7 Operating System using MATLAB R2013b. The algorithm is tested on a variety of indoor and outdoor video sequences of the benchmark IEEE CHANGE DETECTION 2014 dataset [32]. The results for BUNGALOWS, BUSSTATION, COPYMACHINE and CUBICLE dataset are shown in Fig. 4-7 respectively. Where, the column (a) represents moving blob generated by the motion segmentation algorithm, (b) The result of hole filling algorithm, (c) Dual-map representation of the Eigen values for cast and self shadow pixels and (d) Superimposed dual-map on the blob. Based on qualitative analysis of the results, the proposed algorithm satisfactorily segments moving objects, cast and self shadows.

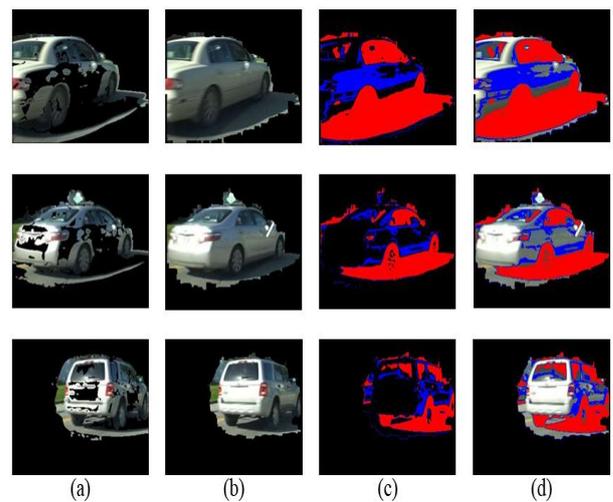

(a)      (b)      (c)      (d)

**Fig 4: Frames in000136, in000247 and in000364 of CHANGE DETECTION 2014 BUNGALOWS dataset results**

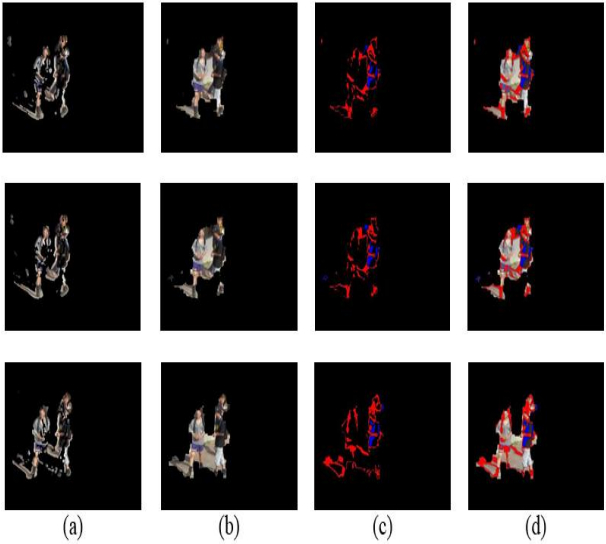

**Fig 5: Frames in000989, in000990 and in000994 of CHANGE DETECTION 2014 BUSSTATION dataset results**

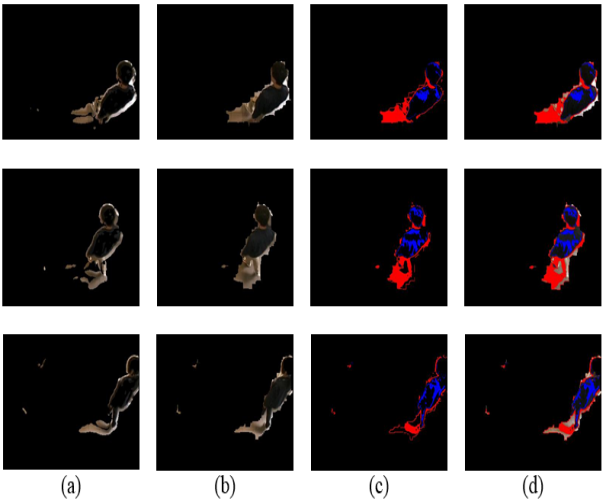

**Fig 6: Frames in000089, in000100 and in000212 of CHANGE DETECTION 2014 COPYMACHINE dataset results**

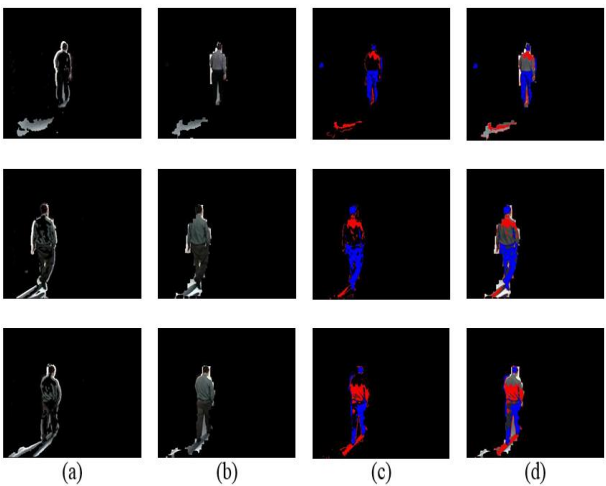

**Fig 7: Frames in001214, in001612 and in001624 of CHANGE DETECTION 2014 CUBICLE dataset results**

In Fig. 4(d), the area under the car is completely detected as cast shadow (red pixel region) because the asphalt on the road and the tyre color of car has similar intensities as the cast shadows. Therefore, while fixing the range of the threshold, the summed Eigen values of entire area will fall within this range. Hence, the area under the car will be segmented. This analysis is applicable for all the results of the dataset i.e., during the camouflage situation the object blob's area even though not completely cast by shadows will be detected as cast shadow and segmented.

Fig. 5 is an example of outdoor video sequence in which the floor color is similar to the skin color of humans. The results in Fig. 5(d) indicates that the cast and self shadows are rightly annotated. However, as discussed earlier misclassification occurs during the camouflage situation.

Fig. 6 and Fig. 7 are examples of indoor sequences with different floor textures and color. The self and cast shadows are detected satisfactorily as shown in Fig. 6(d) and Fig. 7(d). In Fig. 7(d) the direction of light is opposite to the moving object. Hence, the region of the back of object is annotated as self shadow (blue pixel regions).

Identifying self shadows is a very challenging task. An attempt is made in this work to satisfactorily detect self shadow pixels in the object blob without using any background information of the frame.

In order to evaluate the proposed method quantitatively, F-Score as a measure of accuracy is used. The F-Score is the weighted average of precision and recall and is interpreted as accurate if F-Score value reaches 1 [33,34].

$$F = 2 \times \frac{Precision \ \times Recall}{Precision \ + Recall} \qquad (9)$$

Where, $F$ is shadow detection accuracy, $Precision$ is the fraction of retrieved instances that are relevant, and $Recall$ is the fraction of relevant instances that are retrieved.

The evaluation results are shown in Table 1 for sample of results for the IEEE CHANGE DETECTION 2014 dataset. The comparison of result is done with manually segmented ground truth images for the cast and self shadow regions. The mean of F-Score for the Cast and Self shadow is 0.65 and 0.56 respectively. Since, the algorithm is tested on the challenging surveillance video sequences the accuracy of the results of algorithm is above average. Moreover, accurately marking the regions of cast and self shadows manually to generate the ground truth images is a difficult task.

The F-score of self shadow for COPYMACHINE dataset is low compared to the other dataset. This dataset contains humans in dark color clothing which makes it difficult to detect self shadows as the shadow has got the same intensity values as darker regions.

**Table1. Quantitative results of the proposed method**

| *Dataset* | **F** *Cast shadow* | **F** *Self shadow* |
|---|---|---|
| BUNGALOWS | 0.75 | 0.51 |
| BUSSTATION | 0.58 | 0.59 |
| COPYMACHINE | 0.85 | 0.41 |
| CUBICLE | 0.45 | 0.72 |
| ***Mean*** | **0.65** | **0.56** |

The proposed shadows segmentation algorithm is using only the information generated from the object blob. Therefore, in general the computing cost of algorithm is number of pixels in the blob times the computing cost of 3 x 3 neighborhood Eigen values.

## 6. CONCLUSIONS

In this paper, a new method for motion segmentation and shadow segmentation in surveillance videos is proposed. Motion segmentation is done by computing the Euclidian distance between two pixels considering neighborhood values in temporal domain followed by a hole filling post processing technique. The proposed shadow detection method detects cast and self shadows by the dual map interval based representation of the summed Eigen values computed using the neighborhood intensities. The algorithm devised is tested on the challenging Change Detection 2014 Shadow dataset. The results show that the methods satisfactorily segment the moving object, cast and self shadows. Furthermore, the method is also evaluated with shadow detection accuracy metric and the results obtained are promising. In future, adaptive measures for threshold will be considered.

## 7. REFERENCES

[1] Zhang, D. and Lu, G. (2001) Segmentation of Moving Objects in Image Sequence: A Review, Circuits, Systems and Signal Process, vol. 20(2), pp. 143-183.

[2] Prati, A., Cucchiara, R., Mikic I. and Trivedi, M. (2001) Analysis and Detection of Shadows in Video Streams: A Comparative Evaluation, IEEE Trans.

[3] Sanin, A., Sanderson, C. and Lovell, B.C. (2012) Shadow detection: A survey and comparative evaluation of recent methods, ELSEVIER, Pattern Recognition, vol. 45, pp. 1684–1695.

[4] Ullah, H., Ullah, M., Uzair M. and Rehman, F. (2010) Comparative Study: The Evaluation of Shadow Detection Methods, International Journal of Video & Image Processing and Network Security, vol. 01, no. 10.

[5] Heikkila J. and Silven, O. (1991) A real-time system for monitoring of cyclists and pedestrians, Second IEEE Workshop on Visual Surveillance Fort Collins, Colorado, pp. 74-81.

[6] Armanfard, N., Komeili, M. and Kabir, E. (2012) TED: A texture-edge descriptor for pedestrian detection in video sequences, ELSEVIER, Pattern Recognition, vol. 45, pp. 983–992.

[7] Elgammal, A., Duraiswami, R. Harwood, D. and Davis, L.S. (2002) Background and Foreground Modeling Using Nonparametric Kernel Density Estimation for Visual Surveillance, Proceedings of the IEEE, vol. 90(7).

[8] Johnsen S. and Tews, A. (2009) Real-Time Object Tracking and Classification Using a Static Camera, Proceedings of the IEEE ICRA.

[9] Rafael Munoz-Salinas, (2008) A Bayesian plan-view map based approach for multiple-person detection and tracking, ELSEVIER, Pattern Recognition, vol. 41, pp. 3665–3676.

[10] Stauffer, C. and Grimson, W.E.L. (1999) Adaptive background *mixture* models for real-time tracking, Proc. IEEE CVPR 1999, pp. 24&252.

[11] Chao-Yang Lee, Shou-Jen Lin, Chen-Wei Lee and Chu-Sing Yang. (2012) An efficient continuous tracking system in real-time surveillance application, ELSEVIER, Journal of Network and Computer Applications, vol. 35, pp. 1067–1073.

[12] Fang-Hsuan Cheng and Yu-Liang Chen, (2006) Real time multiple objects tracking and identification based on discrete wavelet transform, ELSEVIER, Pattern Recognition, vol. 39, pp. 1126-1139.

[13] Jung-Ho Ahn, Choi, C., Kwak, S. Kim, K. and Byun, H. (2009) Human tracking and silhouette extraction for human-robot interaction systems, Springer-Verlag, Pattern Anal. Appl., vol 12(2), pp 167-177.

[14] Denman, S., Fookes, C. and Sridharan, S. (2010) Group Segmentation During Object Tracking using Optical Flow Discontinuities, Proceedings of IEEE Image and Video Technology (PSIVT), pp.270,275, 14-17.

[15] Chandrajit, M., Girisha, R., and Vasudev, T. (2014a) Motion segmentation from surveillance videos using T-test statistics. In Proceedings of the 7th ACM India Computing Conference (COMPUTE '14). ACM, New York, NY, USA, Article 2, 10 pages. DOI=10.1145/2675744.2675748 http://doi.acm.org/10.1145/2675744.2675748.

[16] Chandrajit, M., Girisha, R., Vasudev, T. (2014b) "Motion Segmentation from Surveillance Video Sequences using Chi-Square Statistics", Proceedings of the second International Conference on - Emerging Research in Computing, Information, Communication and Applications , ERCICA 2014, (Vol 2) 365-372, Elsevier, ISBN 9789351072621.

[17] Girisha R. and Murali, S. (2009) Segmentation of Motion Objects from Surveillance Video Sequences Using Temporal Differencing Combined with Multiple Correlation, Advanced Video and Signal Based Surveillance. Proc. IEEE AVSS '09, pp.472,477, 2-4.

[18] Lim, T., Han B., and Han, J. H. (2012) Modeling and segmentation of floating foreground and background in videos, ELSEVIER, Pattern Recognition, vol. 45, pp. 1696–1706.

[19] Liu, C., Yuen P.C., and Qiu, G. (2009) Object motion detection using information theoretic spatio-temporal saliency, ELSEVIER, Pattern Recognition, vol. 42, pp. 2897-2906.

[20] Lopez, M.T., Fernandez-Caballero, A., Fernandez, M.A., Mira J. and Delgado, A.E. (2006) Visual surveillance by dynamic visual attention method, ELSEVIER, Pattern Recognition, vol. 39, pp. 2194–2211.

[21] Yu-Ting Chen, Chu-Song Chen, Chun-Rong Huang and Yi-Ping Hung, (2007) Efficient hierarchical method for background subtraction, ELSEVIER, Pattern Recognition, vol. 40, pp. 2706–2715.

[22] Amato, A., Mozerov, M.G., Bagdanov, A.D. and Gonzalez, J. (2011) "Accurate Moving Cast Shadow Suppression Based on Local Color Constancy Detection," Image Processing, IEEE Transactions on , vol.20, no.10, pp.2954,2966.

[23] Girisha, R. and Murali, S. (2009) Adaptive Cast Shadow Elimination Algorithm for Surveillance Videos Using t Random Values, Proceedings of India Conference.

[24] Girisha, R. and Murali, S. (2010) Self shadow elimination algorithm for surveillance videos using ANOVA F test, Proceedings of the Third Annual ACM Bangalore Conference.

[25] Jia-Bin Huang, Chu-Song Chen, (2009) "Moving cast shadow detection using physics-based features," Computer Vision and Pattern Recognition, 2009. CVPR 2009. IEEE Conference on , vol., no., pp.2310,2317, 20-25.

[26] Leone, A. and Distante, C. (2007) Shadow detection for moving objects based on texture analysis, ELSEVIER, Pattern Recognition, vol. 40, pp. 1222–1233.

[27] Souza, T., Schnitman, L. and Oliveira, L. (2012) EIGEN ANALYSIS AND GRAY ALIGNMENT FOR SHADOW DETECTION APPLIED TO URBAN SCENE IMAGES, In: IEEE International Conference on Intelligent Robots and Systems (IROS) Workshop on Planning, Perception and Navigation for Intelligent Vehicles.

[28] Chia-Chih Chen; Aggarwal, J.K. (2010) "Human Shadow Removal with Unknown Light Source," Pattern Recognition (ICPR), 2010 20th International Conference on , vol., no., pp.2407,2410, 23-26.

[29] Nadimi, S., Bhanu, B. (2004) "Physical models for moving shadow and object detection in video," Pattern Analysis and Machine Intelligence, IEEE Transactions on, vol.26, no.8, pp.1079,1087.

[30] Wang, Y., Tan, T., Kia-FockLoe and Jian-Kang Wu. (2005) A probabilistic approach for foreground and shadow segmentation in monocular image sequences, ELSEVIER, Pattern Recognition, vol. 38, pp. 1937–1946.

[31] Soille, P. (1999) Morphological Image Analysis: Principles and Applications, Springer-Verlag, pp. 173-174.

[32] Goyette, Jodoin, P.M., Porikli, F., Konrad J. and Ishwar, P. (2012). changedetection.net: A new change detection benchmark dataset, in Proc. IEEE Workshop on Change Detection (CDW-2012) at CVPR-2012, Providence, RI, 16–21.

[33] Powers, D. M. W. (2007) 'Evaluation: From Precision, Recall and F-Factor to ROC, Informedness, Markedness & Correlation' (SIE-07-001), Technical report, School of Informatics and Engineering, Flinders University, Adelaide, Australia.

[34] Van Rijsbergen, C. J. (1979), Information Retrieval (2nd Ed.). Butterworth.